# YOLO Object Detectors for Robotics – a Comparative Study

Patryk Niżeniec, Marcin Iwanowski, Marcin Gahbler
Nicolaus Copernicus University in Toruń, Faculty of Physics, Astronomy and Informatics, Institute of Engineering and Technology,
ul. Grudziądzka 5, 87-100 Toruń, Poland

**Abstract:** YOLO object detectors recently became a key component of vision systems in many domains. The family of available YOLO models consists of multiple versions, each in various variants. The research reported in this paper aims to validate the applicability of members of this family to detect objects located within the robot workspace. In our experiments, we used our custom dataset and the COCO2017 dataset. To test the robustness of investigated detectors, the images of these datasets were subject to distortions. The results of our experiments, including variations of training/testing configurations and models, may support the choice of the appropriate YOLO version for robotic vision tasks.

## 1. Introduction

Contemporary vision systems play a crucial role in robotics, enabling robots to interpret and interact with their environment. Robots can use image processing technologies to analyze visual data, classify objects, determine their location, and make real-time decisions. Vision systems are an integral part of modern robotics and are widely used in its various fields. The vision task, which plays a primary role in robotics, is object detection. Objects are found in a robot workplace, classified into predefined classes, and arranged by bounding boxes.

The rapid development of object detection methods turned them from experimental algorithms into fully functional universal software tools. What motivates their constant development is their vast application possibilities. The number of computer vision tasks that require identifying objects present within a vision scene is comprehensive and covers practically all fields of computer vision. High precision and operational reliability are essential in applications that utilize vision systems. Object detection is crucial in robotics, enabling robots to perceive and interact with their surroundings in real time.

The You Only Look Once (YOLO) family has become popular among the various object detection models due to its high speed and accuracy. YOLO models are advanced single-stage object detection algorithms that do not separate object classification and localization processes into distinct stages. Instead, the algorithm simultaneously analyzes the entire image, dividing it into a grid where each cell predicts bounding boxes and object classes. Thanks to this integrated approach, YOLO is characterized by exceptionally high processing speed, enabling real-time detection and making it a desirable solution for demanding applications. Since its introduction, the application of YOLO models has covered multiple robotic domains, including autonomous navigation, robotic grasping, industrial automation, human-robot interaction (HRI) and aerial robotics. The advancements from YOLOv1 to the latest versions have continuously improved performance, making YOLO an essential tool for real-time robotic perception.

The variety of off-the-shelf object detectors of the YOLO family motivated us to investigate the issue of the appropriate choice of the model version to perform typical robotic tasks, i.e., identification of objects present in the working field of the robotic arm of the manipulator robot. Our investigation focused on popular YOLO models, specifically versions 5, 8, 9, 10, and 11 from Ultralytics. The custom database was prepared to perform tests consisting of images exhibiting various tools and other objects in the workplace. To validate the models' robustness, we introduced a variety of disturbances of images. In our multi-criterion comparison of models, we also considered the computational efficiency of models. Our research results may help find the appropriate YOLO detector model for robotics tasks.

The paper is organized as follows. Section 2 presents the details of the YOLO models. In Section 3, the experimental setup is presented. Section 4 shows the results and main findings of the research. Finally, section 5 concludes the paper.

## 2. Family of YOLO object detectors

The YOLO (You Only Look Once), initially introduced by Joseph Redmon in 2016, YOLO revolutionized object detection by treating it as a single regression problem, allowing the model to predict object locations and classes in a single forward pass through a neural network. The YOLO family of models achieved spectacular success in computer vision. It was, however, preceded by earlier deep learning-based object detectors, such as R-CNN [4], Fast R-CNN [3], and Faster R-CNN [9], which, however, were too slow for real-time deployment in robotics. The introduction of YOLO [8] significantly improved inference speed, allowing real-time obstacle detection in autonomous cars, drones, and robotic delivery systems. Recent versions have been optimized for embedded processors like NVIDIA Jetson, enabling real-time object detection with minimal computational resources.

YOLO detectors are widely used in various robotics domains. Integrating YOLO with depth cameras and robotic vision systems has significantly improved robotic pick-and-place tasks [13]. In logistics and e-commerce warehouses, YOLO-based grasping systems enable robots to recognize, pick, and sort items efficiently. The introduction of YOLO-based anomaly detection has enabled real-time inspection of products on assembly lines [7]. YOLO models trained on industrial datasets can rapidly detect scratches, cracks, misalignments, and other defects, enhancing quality control processes in real-time [16]. In healthcare robotics, patient monitoring and assistive robots use YOLO to recognize human postures and facial expressions, allowing better interaction with users [19]. Social robots and service robots in public spaces rely on YOLO for real-time person tracking, enabling gesture recognition, emotion detection, and activity recognition, making it more effective for robotic assistants in smart environments [13]. In precision agriculture, drones utilize YOLOv8 for plant health monitoring, pest detection, and yield estimation, helping farmers optimize crop production [21].

Over the years, the YOLO family has evolved through multiple versions, each improving upon the previous in accuracy, efficiency, and ease of use. The early versions: YOLOv1 [8], YOLOv2 [7] and YOLOv3 [19] focused on balancing speed and accuracy, while YOLOv4 [16] significantly improved training techniques and optimizations. YOLOv5 [20], maintained by Ultralytics, further refined the model by implementing it in PyTorch, making it more accessible and lightweight. Subsequent versions, including YOLOv6 [18], YOLOv7 [13], and YOLOv8 [21], introduced additional architectural enhancements, training strategies, and deployment efficiencies, catering to different industry needs. YOLOv9 introduces Programmable Gradient Information (PGI) and Generalized Efficient Layer Aggregation Network (GELAN) technologies, which significantly enhance the object detection process. YOLOv10 brings key improvements by eliminating the need for NMS through Dual Label Assignments. YOLO11[1] is based on the YOLOv8 architecture, introducing significant enhancements that improve both performance and detection accuracy. The main change is the replacement of the C2f block with the new C3k2, which speeds up processing by using two smaller convolutions. Principal properties of consecutive YOLO versions are listed in Table 1.

In our research, we focused on YOLO models released by Ultralytics, i.e., versions 5 [20], 8 [21], 9 [14], 10 [12], and 11 [22], which became one of the most popular tools for object detection due to its ease of use and robust implementations[2].

To accommodate diverse computational requirements, YOLO models, beginning with version 5, introduced systematic model variants. The most lightweight variants (e.g., YOLOv5n) are optimized for real-time processing on edge devices, while mid-sized variants (e.g., YOLOv5s/m) balance speed and accuracy. Larger variants (e.g., YOLOv5l/x) prioritize higher accuracy at the cost of increased computational demands, making them suitable for powerful GPUs and high precision applications. This tiered approach allows users to select models tailored to their hardware constraints and detection needs. Subsequent YOLO versions (e.g., YOLOv6, v7, v8) have adopted and refined this strategy, ensuring compatibility with evolving edge-to-cloud workflows. Various YOLO models and versions were compared in a few papers, focusing on structural comparison of models [10, 11], as well as on particular applications in agriculture [15], object recognition [2], real-time vision [17], remote sensing [1] and industry [5].

---

[1] Ultralytics has simplified the naming convention for YOLO11 by removing the 'v' that was present in previous versions of YOLO

[2] Missing versions: 6 and 7 were omitted as they are not part of the Ultralytics library. Version 6, though technically implemented, lacks COCO-trained weights critical for comparison.

**Table 1. Comparison of YOLO Versions 1 to 11 (versions tested in the current study in bold)**
Tabela 1. Porównanie wersji YOLO od 1 do 11 (wersje przetestowane w niniejszym badaniu wyróżniono pogrubieniem)

| Version | Variants | Released | Architecture | Key Features |
|---|---|---|---|---|
| v1 | standard | June 2016 | Custom CNN | Single-stage, real-time detection |
| v2 | standard | December 2016 | Darknet-19, Anchor Boxes | Anchor Boxes, BatchNorm, Multi-label |
| v3 | tiny, standard, large | April 2018 | Darknet-53, Multi-scale Predictions | Deeper network, Multi-scale detection |
| v4 | small, medium, large, extra large | April 2020 | CSPDarknet53, PANet/FPN | Optimized training, Better feature fusion |
| **v5** | **nano, small, medium, large, extra large** | **June 2020** | **CSPDarknet53, PANet/FPN** | **PyTorch-based, AutoAnchor** |
| v6 | nano, small, medium, large | June 2022 | EfficientRep Backbone, RepVGG Neck | SimOTA, TensorRT-Optimized |
| v7 | standard, extra large | July 2022 | Extended ELAN, Re-param Conv | Task-specific, Auxiliary Head |
| **v8** | **nano, small, medium, large, extra large** | **January 2023** | **CSP-Inspired, Anchor-Free Head** | **Adaptive Tuning, No Anchors** |
| **v9** | **tiny, small, medium, custom, extended** | **February 2024** | **GELAN, PGI** | **Gradient-Preserving, Efficient** |
| **v10** | **nano, small, medium, balanced, large, extra large** | **May 2024** | **SC Decoupled Downsampling** | **No-NMS, Edge-Optimized** |
| **11** | **nano, small, medium, large, extra large** | **October 2024** | **Lightweight Backbone** | **Feature Fusion, Speed-Optimized** |

## 3. Experimental setup

The study aims to analyze the robustness of YOLO models to introduce distortions in both a custom dataset developed for robotics applications and the Microsoft COCO 2017 dataset [6]. To enhance the model's robustness and generalization capability, data augmentation was applied to the training and validation subsets. The applied transformations included noise addition, blurring, brightness adjustment, hue shift, and rotation with scaling.

During model training, only the training and validation subsets were used. The training subset was employed to update the model's weights, while the validation set enabled the selection of the best-trained variant. The test subset was not involved in the training process and was used solely to evaluate the final models' robustness and precision. For testing purposes, distortions analogous to the previous augmentations were added to the test subset of the custom dataset and the validation subset of the Microsoft COCO dataset, but with different parameters. In this case, separate subsets were created, each containing only one type of distortion, allowing for an assessment of the impact of individual distortions on model performance. For the COCO dataset, distortions related to rotation and hue modification were omitted, as they do not reflect the real-world distortions present in this dataset.

### 3.1. Datasets

The research was primarily based on our custom dataset, where images present various tools in the robot workplace. The custom dataset, specifically created for this research, included images of 23 classes of workshop tools commonly used in daily work. These tools included items such as hammers, pliers, files, and wrenches. This dataset was captured in a typical robotic environment, with uniform lighting and a consistent, smooth background, ensuring clear and well-lit images. All images were taken from a single camera perspective, simulating a camera mounted above the robot, which facilitated obtaining consistent and easily analyzable data. The custom dataset was divided into three subsets: a training set containing 439 images, validation comprising 154 images, and a test consisting of 162 images. In addition, the Microsoft COCO 2017 [6] is used. It is divided into three subsets: training containing 118k images, validation with 5k images, and test with 20k images. Experiments performed on two datasets provide a broader foundation for evaluating the effectiveness of the proposed object detection methods. Example images from both datasets are shown in Figure 1.

Data augmentation techniques were applied to both training and validation subsets to enhance the model's robustness and improve its generalization capabilities. Augmentation involves artificially expanding the dataset through various transformations, increasing its diversity, and mitigating the risk of overfitting and enhancing the model's performance in real-world conditions.

As part of the augmentation process, the following transformations were introduced:

– Noise Addition: Introduces "salt and pepper" noise into the image by generating two random matrices representing white (salt) and black (pepper) pixels. The noise intensity is controlled by a specified probability parameter.
– Blurring: Applies a Gaussian blur effect to the image using OpenCV's `GaussianBlur` function. The strength of the blur is determined by the size of the filter kernel, which is adjustable via a user-defined parameter.
– Brightness Adjustment: Modifies the image brightness by converting the image to the HSV colour space and scaling the "Value" channel by a specified percentage. Pixel values are clipped to ensure they remain within the valid range of 0 to 255 before converting the image back to the BGR color space.
– Hue Shift: Alters the image's hue by converting it to the HSV color space and adding a predefined offset to the "Hue" channel. This adjustment allows for precise control over the color tone of the image.
– Rotation and Scaling: Rotates the image by a specified angle and scales it to fit within the original dimensions. The rotation is performed using the `warpAffine` function, and the final image is seamlessly blended with a background using seamless cloning to ensure a natural appearance.

These transformations contributed to a more diverse dataset, enabling the model to more effectively recognize objects under conditions differing from those in the original dataset. Parameters of used augmentations are shown in Table 2. As a result, the number of images in the training subset increased to 12 292, while the validation subset expanded to 4 312. Figure 2 illustrates sample images after augmentation.

Additionally, to conduct more rigorous tests of the model's resilience to various types of distortions, further transformations were applied to the test subset of the custom dataset and the validation subset of the Microsoft COCO dataset. These distortions were analogous to the previously employed augmentations but were adjusted to intensify the degree of disturbance compared to those in the training and validation sets. Used param-

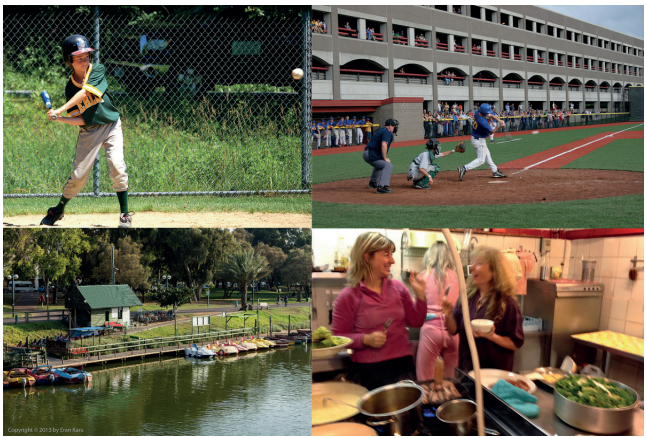
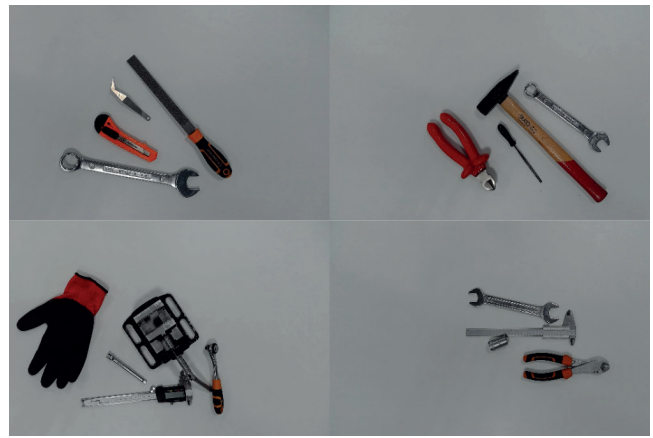

**Fig. 1. Example images from the COCO2017 (left) and from the custom dataset (right)**
Rys. 1. Przykładowe obrazy ze zbioru COCO2017 (po lewej) oraz z autorskiego zestawu danych (po prawej)

**Table 2. Parameters of the augmentation techniques applied to the training and validation subsets**
Tabela 2. Parametry technik augmentacji zastosowanych na podzbiorach treningowych i walidacyjnych

|  | Lower Bound | Upper Bound | Step |
|---|---|---|---|
| Noise addition | 5 | 17 | 3 |
| Blurring | 5 | 14 | 3 |
| Brightness adjustment | 75 % | 120 % | 9 % |
| Hue shift | 15° | 180° | 33° |
| Rotation and scaling | 15° | 147° | 33° |

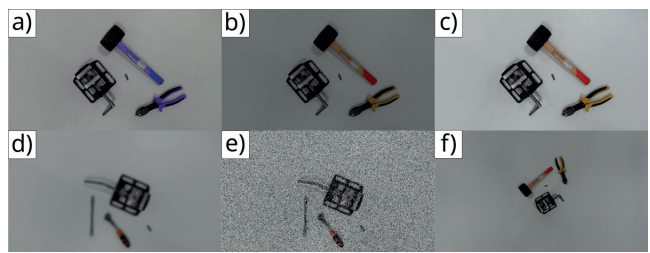

**Fig. 2. Example images with added augmentation: a) hue shift, b) darkening, c) brightening, d) blurring, e) noise addition, f) rotation and scaling**
Rys. 2. Przykładowe obrazy z zastosowaną augmentacją: a) zmiana odcienia, b) przyciemnienie, c) rozjaśnienie, d) rozmycie, e) dodanie szumu, f) obrót i skalowanie

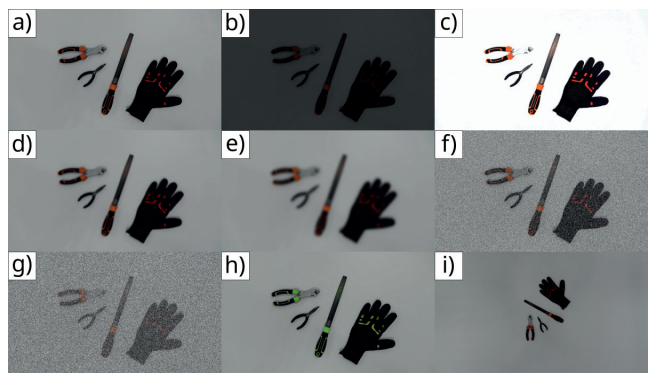

**Fig. 3. Example images with added distortions: a) no distortions, b) darkening, c) brightening, d) mild blurring, e) strong blurring, f) mild noise addition, g) strong noise addition, h) hue shift, i) rotation and scaling**
Rys. 3. Przykładowe obrazy z dodanymi zakłóceniami: a) brak zakłóceń, b) przyciemnienie, c) rozjaśnienie, d) lekkie rozmycie, e) silne rozmycie, f) dodanie lekkiego szumu, g) dodanie silnego szumu, h) zmiana odcienia, i) obrót i skalowanie eters are shown in Table 3. To precisely assess the impact of specific distortions on model performance, dedicated test subsets were created, each containing images modified by a single type of transformation. Importantly, all distortions were introduced at the original resolution before any resizing was applied. This approach ensured that the distortions closely resembled real-world conditions, where image degradation typically occurs before rescaling. Example images after adding distortions are shown in Figure 3.

**Table 3. Parameters of distortions applied to the test subsets**
Tabela 3. Parametry zakłóceń zastosowanych na podzbiorach testowych

|  | Lower Bound | Upper Bound | Step |
|---|---|---|---|
| Mild noise addition | 15 | 35 | 4 |
| Strong noise addition | 40 | 60 | 4 |
| Mild blurring | 5 | 35 | 6 |
| Strong blurring | 35 | 65 | 6 |
| Brightening | 200 % | 400 % | 40 % |
| Darkening | 10 % | 50 % | 8 |
| Hue shift | 15° | 345° | 66° |
| Rotation and scaling | 15° | 345° | 66° |

For the COCO dataset, distortions related to rotation and hue modification were omitted, as they do not reflect typical disturbances that might occur in this dataset. This decision ensured that the testing conditions remained aligned with real-world scenarios in which object detection models are deployed.

### 3.2. Experiments workflow

A structured testing methodology was implemented to ensure a thorough evaluation of the YOLO models. The primary objective was to assess the models' robustness to various distortions by analyzing their performance on pre-prepared test subsets subjected to controlled distortions.

Object detection models require constant image sizes determined by the size of the detector input layer. Since original images may have various sizes, the resampling needs to be applied to match the original size with one of the input layers. Moreover, the pixel values are normalized and scaled. Most YOLO models perform the resizing automatically using the bilinear interpolation. It is possible, however, to perform this rescaling using alternative methods, simply including this resampling in the processing pipeline just before the detector. In our experiments, we apply an alternative resampling – the area interpolation, which was used in YOLOv6, the model that does not belong to the Unltralytics zoo [18].

The evaluation of model precision was conducted using the standard mAP(50–95) metric, which measures Mean Average Precision across multiple IoU (Intersection over Union) thresholds ranging from 50 % to 95 %. This metric comprehensively assesses detection accuracy across varying degrees of overlap between predicted and ground-truth bounding boxes.

To quantify the impact of image distortions, a robustness metric was introduced. Robustness was defined as the difference between the precision achieved on the undistorted test subset and the performance on subsets containing specific distortions. The overall robustness of a given model was computed as the average robustness across all tested distortions.

Additionally, analogous experiments were conducted using models fine-tuned on the Microsoft COCO 2017 dataset. As in previous evaluations, distortions related to hue shift and rotation with scaling were excluded due to their limited relevance in real-world COCO dataset conditions. Hue modifications could introduce unnatural color changes (e.g., unrealistic skin tones in human images), while rotation and scaling could lead to loss of image quality, potentially skewing the evaluation results. Tests were conducted to assess the image analysis performance of YOLO models and determine the analysis time for a single image in various configurations. The model's speed depends on its size and the device's computational power on which it is deployed. More extensive models offer better detection accuracy but have longer inference times, while smaller models are faster but may exhibit lower precision.

The tests measured image analysis time for different versions of the models, both in PyTorch format and after optimization to the TensorRT format. The goal was to compare inference times and understand how model optimization affected analysis speed while maintaining adequate detection accuracy.

## 4. Results

### 4.1. Model training

The research began with training models on a custom dataset. Models from the Ultralytics library were used, originally pre-trained on the Microsoft COCO 2017 dataset. Most training parameters were left unchanged, with only the number of epochs set to 50. A such number of epochs was the optimal value for the given dataset, balancing training time and performance. The training process and subsequent inference tests were conducted on a workstation running Windows 11, equipped with an NVIDIA GeForce RTX 3060 GPU (12 GB VRAM), ensuring consistent hardware conditions for all experiments. Additionally, the batch size was set to 8, based on the computational capabilities of the available GPU. A plot showing the relationship between the number of model parameters and training time for the augmented dataset is shown in Figure 4. The models were trained on the dataset without augmentation and the dataset with applied augmentation to assess its impact on model performance.

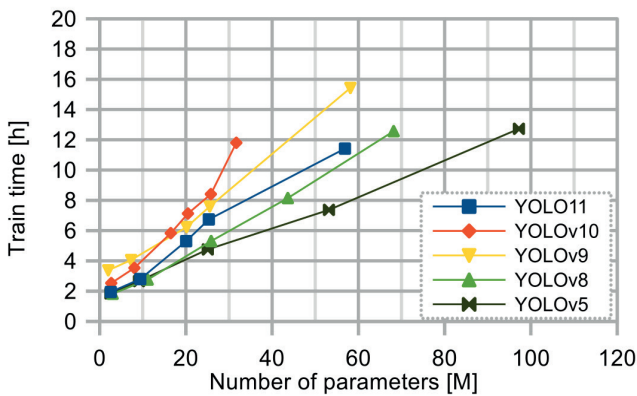

**Fig. 4. Relationships between the number of YOLO model parameters and their training time on the augmented dataset. Distinct points on each curve correspond to consecutive model variants (e.g., nano, small, medium, or version-specific equivalents) in ascending order of complexity**
Rys. 4. Zależności między liczbą parametrów modeli YOLO a czasem ich treningu na augmentowanym zbiorze danych. Poszczególne punkty na każdej krzywej odpowiadają kolejnym wariantom modeli (np. nano, small, medium lub ich odpowiednikom specyficznym dla danej wersji) uporządkowanym według rosnącej złożoności

**Table 4. Performance of YOLO models trained with and without augmentation on undisturbed data (mAP 50–95)**
Tabela 4. Wydajność modeli YOLO trenowanych z augmentacją i bez augmentacji na zbiorze bez zakłóceń (mAP 50–95)

| Model | Augmented | Original |
|---|---|---|
| YOLO11 | 80.46 % | 61.75 % |
| YOLOv10 | 80.07 % | 57.50 % |
| YOLOv9 | 81.80 % | 63.68 % |
| YOLOv8 | 81.20 % | 61.09 % |
| YOLOv5 | 81.07 % | 61.40 % |

Training was performed for selected variants of the YOLO models of different variants. Subsequently, tests were conducted on the previously prepared test subsets. To compare the effectiveness of both training data sets, the mean average precision (mAP) was calculated for all model sizes tested, and the results are presented as an average for the particular model versions in Table 4. This first experiment was based on undistorted images. Models trained on the augmented dataset achieved significantly better results than those trained without augmentation. Owing to the latter, all the subsequent experiments employed models trained using the augmented data.

As indicated in Table 4, the application of data augmentation resulted in a substantial performance increase across all models, with mAP improvements ranging from approximately 18 to 20 percentage points compared to models trained on the original dataset.

Furthermore, Table 5 demonstrates that for noisy environments, area-based interpolation significantly outperforms bilinear interpolation, preserving detection capabilities where the standard approach fails (e.g., YOLOv8 achieves 78 % vs 29 %).

### 4.2. Influence of interpolation method

Image interpolation, used to match the size of the input image size with the size required by the detector model, can significantly impact the performance of object detection models, especially in noisy environments. To evaluate the effect of different interpolation methods, the mean average precision (mAP) was calculated by testing bilinear and area-based interpolation on a test subset with strong noise addition. The results were aggregated and presented as averages for each model version, enabling a comprehensive comparison of their performance. The findings for each method are summarized in Table 5.

A key factor in this evaluation was the order of preprocessing steps. This study applied distortions such as noise before interpolation, reflecting real-world scenarios where noise is often introduced at the original resolution before any rescaling occurs. This approach ensured that the interpolation process interacted with the already degraded image, which can influence how distortions are preserved or altered.

**Table 5. Performance of YOLO models under strong noise with different interpolation methods (mAP 50–95)**
Tabela 5. Wydajność modeli YOLO przy silnym szumie z różnymi metodami interpolacji (mAP 50–95)

| Model | Area-based Interpolation | Bilinear Interpolation |
|---|---|---|
| YOLO11 | 74.57 % | 21.82 % |
| YOLOv10 | 75.57 % | 28.00 % |
| YOLOv9 | 78.89 % | 28.49 % |
| YOLOv8 | 78.28 % | 29.44 % |
| YOLOv5 | 79.48 % | 27.68 % |

The results confirmed that area-based interpolation offers superior performance. As demonstrated in Table 5, for noisy environments, area-based interpolation significantly outperforms bilinear interpolation, preserving detection capabilities where the standard approach fails (e.g., YOLOv8 achieves ~78 % vs ~29 %). Area-based interpolation appears to better preserve essential structural details while reducing the impact of high-frequency noise artifacts. Further research is being conducted on datasets with area-based interpolation applied as part of the data preprocessing to understand its advantages in various conditions better.

### 4.3. Influence of image distortions

As part of the conducted research, the precision of all analyzed models was assessed using preprepared datasets containing distortions. The study covered models fine-tuned on both the custom and Microsoft COCO 2017 datasets, enabling a comprehensive analysis of their performance under various test conditions. The results for the custom dataset are presented in Figure 5, while the results for the COCO 2017 dataset are shown in Figure 6. Colors in all charts refer to particular detector YOLO version, while marks refer to model variant.

Model testing on the custom dataset revealed minimal performance differences. For the test subset without disturbances, the difference in mAP scores between the best and worst-performing models was only 2.26 percentage points. This slight discrepancy suggests that, in this type of task, the choice of a specific model does not significantly affect the results obtained.

The models were also tested on the COCO dataset, where the differences between the best and worst-performing models were more significant, amounting to 21.03 percentage points for the subset without disturbances. This is due to the greater complexity of the COCO dataset and the greater diversity of images, which can lead to differences in detection precision between models.

The models demonstrated the greatest difficulties in the presence of intense noise (Figure 5g) and brightness variations (Figure 5b,c) in the custom dataset, as well as strong blur (Figure 6e) and noise (Figure 6f, g) in the Microsoft COCO dataset. Regarding the custom dataset, the best results were obtained for images with blur (Figure 5d, e), mild noise (Figure 5f), hue changes (Figure 5h), and transformations involving rotation and scaling (Figure 5i). In contrast, for the Microsoft COCO dataset, the models performed best under brightness variations (Figure 6b, c) and mild blur (Figure 6d).

These differences stem from the characteristics of both datasets and the preprocessing applied. In the custom dataset, noise was added before resizing, using area interpolation, which may have smoothed out some noise artifacts while preserving structural details. This approach was chosen because, in a real-world scenario, noise would naturally be introduced at the original resolution before any resizing occurs. Conversely, in the Microsoft COCO dataset, images were first resized and then noise was applied, meaning that noise was introduced at the final resolution without being affected by interpolation, potentially making it more disruptive. This difference was due to the constraints of working with the pre-existing Microsoft COCO dataset, where direct control over noise introduction before resizing was not possible.

Additionally, the Microsoft COCO dataset features greater diversity and a significantly larger number of images, which enhances the models' robustness to variations in lighting intensity.

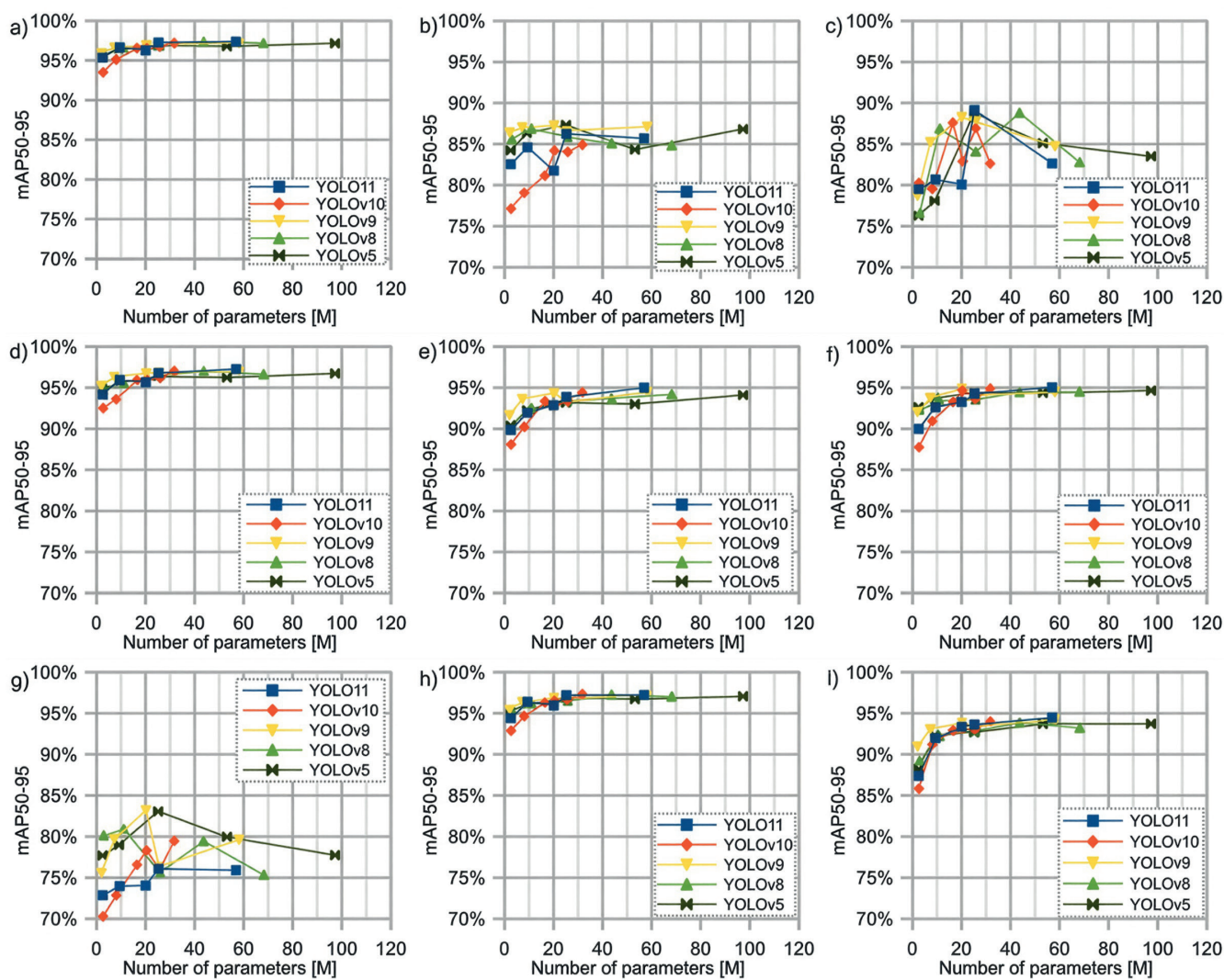

**Fig. 5.** Set of plots presenting the precision test results for models trained on a custom dataset: a) no distortions, b) darkening, c) brightening, d) mild blurring, e) strong blurring, f) mild noise addition, g) strong noise addition, h) hue shift, i) rotation and scaling
Rys. 5. Zbiór wykresów przedstawiających wyniki testów precyzji dla modeli trenowanych na autorskim zbiorze danych: a) brak zakłóceń, b) przyciemnienie, c) rozjaśnienie, d) lekkie rozmycie, e) silne rozmycie, f) dodanie lekkiego szumu, g) dodanie silnego szumu, h) zmiana odcienia, i) obrót i skalowanie

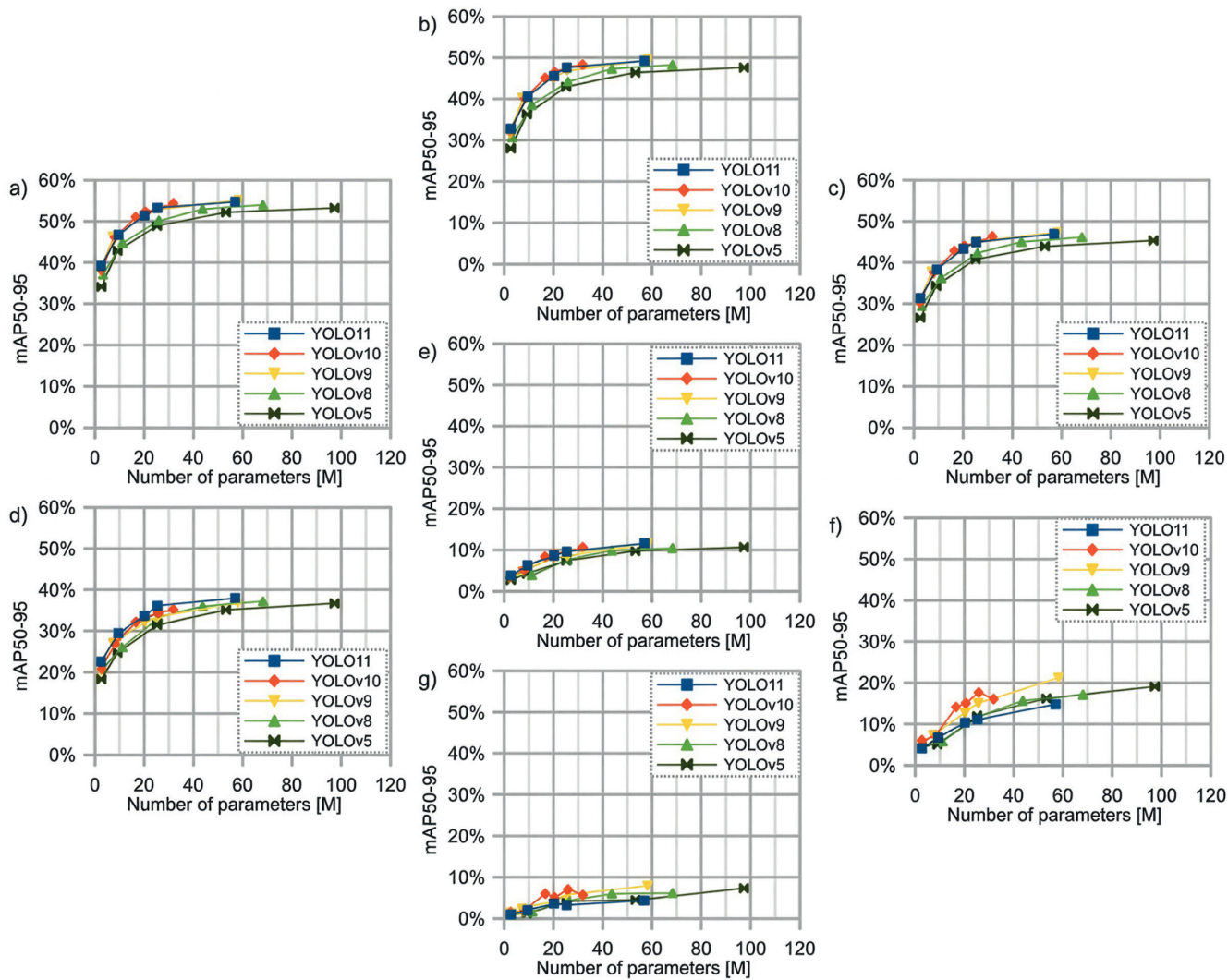

**Fig. 6. Set of plots presenting the precision test results for models trained on a Microsoft COCO dataset: a) no distortions, b) darkening, c) brightening, d) mild blurring, e) strong blurring, f) mild noise addition, g) strong noise addition**
Rys. 6. Zbiór wykresów przedstawiających wyniki testów precyzji dla modeli trenowanych na zbiorze Microsoft COCO: a) brak zakłóceń, b) przyciemnienie, c) rozjaśnienie, d) lekkie rozmycie, e) silne rozmycie, f) dodanie lekkiego szumu, g) dodanie silnego szumu

However, models trained on the Microsoft COCO dataset exhibited lower resilience to distortions that lead to the loss of image details, such as noise and strong blur. This suggests that models trained on more diverse datasets adapt better to global changes, such as lighting, but are less resistant to the degradation of local image details.

On the other hand, models trained on the custom dataset handled blur more effectively because, despite image degradation, the structure and color characteristics of the tools were preserved. This can be attributed to prior data augmentation, which increased the model's resistance to such distortions, even when intensified in test datasets. Additionally, since the noise was applied before resizing using area interpolation, some high-frequency noise artifacts were smoothed out in the process, reducing their disruptive impact on the model's performance. As a result, the model retained better feature representation despite noise introduction. Furthermore, object identification remains easier in a less complex images despite modifications, which may contribute to higher detection accuracy under specific conditions.

Based on the obtained precision results, robustness was calculated to assess whether there are differences in the resilience of various YOLO versions to the presence of distortions. This analysis provided valuable insight into how each model responds to disturbances, highlighting their adaptability in real-world applications. Furthermore, the results guide selecting the most robust model for deployment in environments with different distortions. The results are presented in Figure 7.

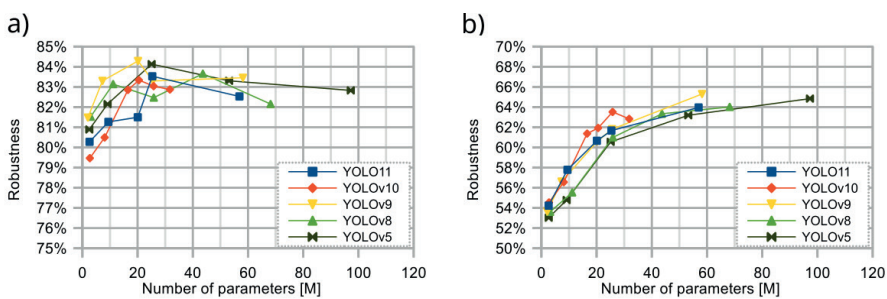

**Fig. 7. Comparison of model robustness to distortions for models fine-tuned on a) the custom dataset and b) the Microsoft COCO 2017 dataset**
Rys. 7. Porównanie odporności modeli na zakłócenia dla modeli douczonych na a) autorskim zestawie danych i b) zbiorze Microsoft COCO 2017

The differences in robustness to distortions between versions of the YOLO models are minimal for both the custom and Microsoft COCO datasets. In the case of models fine-tuned on the COCO dataset, a clear trend was observed where robustness increased with the complexity of the models. However, for models trained on the custom dataset, the smallest models performed worse than those with a larger number of parameters. The reason is that the training data was insufficient for the less complex dataset to effectively train smaller models. Larger-size models achieved the best compromise between fitting and generalization ability, resulting in higher robustness to distortions.

### 4.4. Inference Speed and Deployment Optimization

In object detection, the inference time of a single image is a key factor. One of YOLO's most significant advantages is its ability to operate in real-time; however, the model's speed depends on both its size and the computational power of the device on which it is deployed. Generally, larger models achieve higher detection accuracy but operate significantly slower. Figure 8a presents the measured image analysis times for various models, demonstrating that selecting an appropriate model should consider the specific requirements of the task at hand. Using the RTX 3060 GPU, even the larger models (L/X variants) achieved inference times below 30 ms, as shown in Figure 8a, which qualifies them for real-time robotic applications.

Before deploying a model, testing its performance on the target hardware is recommended. For devices with limited computational power, such as single-board computers like Raspberry Pi or smartphones, smaller models can be used, provided they are properly exported. In contrast, larger models require more powerful computing units, such as PCs equipped with modern GPUs. When accelerating models on GPUs, NVIDIA units are the best choice, as they allow model export to the TensorRT format. This optimization can significantly boost performance up to five times without sacrificing accuracy. However, it is essential to note that such optimization tailors the model specifically to a given device, meaning that a model exported to TensorRT will not function correctly on different hardware. Figure 8b presents performance test results for models optimized in this manner.

Alternative optimization methods such as ONNX Runtime and OpenVINO provide deployment flexibility across different hardware platforms. Techniques like model quantization and pruning can reduce computational demands, making models more efficient for edge devices. However, other optimization methods may introduce compatibility constraints. For example, TensorRT might not support specific custom layers, while ONNX and OpenVINO may have their own limitations depending on the target hardware. Ensuring compatibility and extensively testing the model on the target device is crucial for achieving optimal performance.

## 5. Conclusions

The research described in this paper aimed to investigate the applicability of models from the popular YOLO family to robotic tasks. The results indicate that the differences between the analyzed models are minimal for less complicated images, such as those examined in this study. The gap between the best and worst mAP results in the absence of noise is only 2.26 percentage points. For instance, on the custom dataset, the performance gap between the lightweight YOLOv10n and the complex YOLOv10x was minimal, whereas on COCO it exceeded 20 %. The robustness to noise across different model versions remains similar for the analyzed and Microsoft COCO datasets. In expected distortions, selecting a specific model version is not particularly significant. Models with a small number of parameters show the least robustness, whereas larger models achieve the best results.

Additionally, area interpolation provides better results for noisy images than bilinear interpolation when downscaling. Experiments confirmed that switching to area interpolation can recover nearly 50 percentage points of mAP in strongly noisy conditions (Table 5). Since YOLO automatically adjusts image resolution using bilinear interpolation, which preserves more details but is more susceptible to noise, it is recommended to reduce the resolution beforehand using area interpolation in noisy images. Furthermore, models trained on the Microsoft COCO dataset demonstrate greater robustness to global lighting changes but struggle with strong blur and noise due to the loss of local image details. In contrast, models trained on a custom dataset handle blur more effectively but are less resistant to global lighting variations.

Exporting the model to TensorRT format significantly reduces inference time without affecting precision when using an NVIDIA GPU. This format enables the model to run outside the Python environment while optimizing its performance for the target hardware. It is also observed that large-parameter models do not exhibit significantly higher precision than those with fewer parameters, but they operate considerably slower. Finally, applying data augmentation to the training set significantly improves mAP. These conclusions may serve as guidelines for selecting an appropriate object detection model for robotic applications.

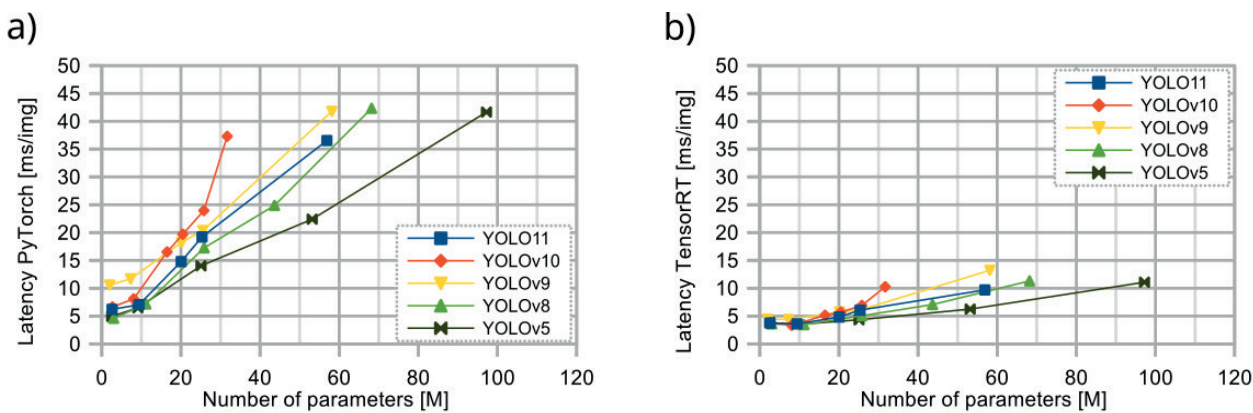

**Fig. 8. Comparison of model inference time for a single image in a) PyTorch format and b) TensorRT format**
Rys. 8. Porównanie czasu inferencji modelu dla pojedynczego obrazu w formacie a) PyTorch oraz b) TensorRT

# Detektory obiektów YOLO w robotyce – analiza porównawcza

Streszczenie: Detektory obiektów YOLO stały się ostatnimi czasy kluczowym elementem systemów wizyjnych w wielu dziedzinach. Rodzina dostępnych modeli YOLO składa się z wielu wersji, z których każda występuje w różnych wariantach. Badania opisane w niniejszej pracy mają na celu zweryfikowanie przydatności członków tej rodziny do wykrywania obiektów znajdujących się w przestrzeni roboczej robota. W eksperymentach wykorzystano nasz własny zbiór danych oraz zbiór COCO2017. Aby przetestować odporność badanych detektorów, obrazy z tych zbiorów poddano zniekształceniom. Wyniki eksperymentów, uwzględniające różne konfiguracje treningowe/testowe oraz modele, mogą stanowić wsparcie przy wyborze odpowiedniej wersji YOLO dla zadań związanych z wizją robotyczną.

**Słowa kluczowe:** percepcja robota, analiza obrazu, sztuczna inteligencja, detektory YOLO, widzenie komputerowe, uczenie głębokie, detekcja obiektów

**Patryk Niżeniec, MSc Eng.**
pnizeniec@umk.pl
ORCID: 0009-0001-1859-9299

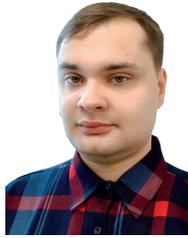

He received his engineering degree in 2023 and master's degree in 2024 in Automation and Robotics from the Nicolaus Copernicus University in Toruń, The Faculty of Physics, Astronomy and Informatics, Institute of Engineering and Technology. Since 2024, he has been employed there as a research and teaching assistant. After completing his MSc degree, he began research work toward a doctoral degree in the field of robotics and computer vision. His research interests include Gaussian Splatting, computer vision for robotic systems, and synthetic dataset generation for machine learning applications. His work focuses on vision-based perception, scene reconstruction, deep neural networks, and the integration of artificial intelligence methods with robotic platforms. This article constitutes his first scientific publication.

**Prof. Marcin Iwanowski, DSc PhD**
iwanowski@fizyka.umk.pl
ORCID: 0000-0001-8347-1112

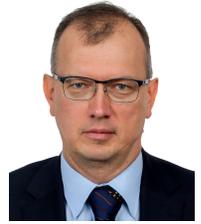

He is a university professor at the Centre for Technical Sciences, Faculty of Physics, Nicolaus Copernicus University in Toruń, and at the Institute of Control and Industrial Electronics, Faculty of Electrical Engineering, Warsaw University of Technology. His research interests focus on computer vision, machine learning, and deep learning methods applied to data analysis, including visual data, time series, and text. He is the author or co-author of more than 100 scientific publications.

**Marcin Gahbler, MSc**
gabi@fizyka.umk.pl
ORCID: 0009-0008-3790-7112

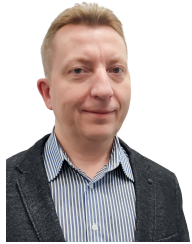

Employee of the Institute of Engineering and Technology at Nicolaus Copernicus University in Toruń. Graduate of the Faculty of Physics, Astronomy and Informatics at Nicolaus Copernicus University, where he obtained a degree in physics, specialising in the physical foundations of microelectronics. Long-time supervisor of workshops and laboratories in the field of industrial automation. Currently involved in the application of vision systems in automation systems.